\documentclass[runningheads]{llncs}

\usepackage{eccv}

\usepackage{eccvabbrv}

\usepackage{graphicx}
\usepackage{booktabs}

\usepackage[accsupp]{axessibility}  
\usepackage[dvipsnames]{xcolor}

\usepackage{hyperref}
\usepackage{multirow}
\usepackage[figuresright]{rotating}
\usepackage{amsmath}
\usepackage{color}

\definecolor{newgreen}{rgb}{0, 0.6, 0.2}

\usepackage{orcidlink}

\begin{document}

\title{MesonGS: Post-training Compression of 3D Gaussians via Efficient Attribute Transformation} 
\titlerunning{MesonGS: Post-training Compression of 3D Gaussians}

\author{Shuzhao Xie\inst{1}\orcidlink{0009-0008-3017-1077} \and
Weixiang Zhang\inst{1} \and
Chen Tang\inst{1,3} \and
Yunpeng Bai\inst{4} \and 
Rongwei Lu\inst{1} \and 
Shijia Ge\inst{1} \and 
Zhi Wang\inst{1,2}$^{,\dagger}$\orcidlink{0000-0002-5462-6178}}

\authorrunning{S.~Xie et al.}

\institute{$^1$ SIGS \& TBSI, Tsinghua University, 
$^2$ Peng Cheng Laboratory \\
$^3$ MMLab, The Chinese University of Hong Kong, 
$^4$ The University of Texas at Austin \\
\url{https://shuzhaoxie.github.io/mesongs/}}

\maketitle

\def\thefootnote{$\dagger$}\footnotetext{Corresponding author.}

\begin{abstract}
    3D Gaussian Splatting demonstrates excellent quality and speed in novel view synthesis. Nevertheless, the huge file size of the 3D Gaussians presents challenges for transmission and storage. Current works design compact models to replace the substantial volume and attributes of 3D Gaussians, along with intensive training to distill information. These endeavors demand considerable training time, presenting formidable hurdles for practical deployment. To this end, we propose \emph{MesonGS}, a codec for post-training compression of 3D Gaussians. Initially, we introduce a measurement criterion that considers both view-dependent and view-independent factors to assess the impact of each Gaussian point on the rendering output, enabling the removal of insignificant points. Subsequently, we decrease the entropy of attributes through two transformations that complement subsequent entropy coding techniques to enhance the file compression rate. More specifically, we first replace rotation quaternions with Euler angles; then, we apply region adaptive hierarchical transform to key attributes to reduce entropy. Lastly, we adopt finer-grained quantization to avoid excessive information loss. Moreover, a well-crafted finetune scheme is devised to restore quality. Extensive experiments demonstrate that MesonGS significantly reduces the size of 3D Gaussians while preserving competitive quality.
    \keywords{Compression \and Gaussian Splatting \and Novel View Synthesis} 
\end{abstract}
    
\section{Introduction}
\label{sec:intro}
Novel view synthesis is a fundamental task in 3D vision and has significant applications in virtual reality, augmented reality, and photography. This task involves using a collection of images captured from different viewpoints, along with their corresponding camera poses, with the objective of generating highly realistic images from arbitrary viewpoints.
By reparameterizing the point with a 3D Gaussian function in the point cloud, 3D Gaussian Splatting (3D-GS) \cite{kerbl20233d} shows excellent quality and real-time rendering speed in this task. A Gaussian point consists of a 3D coordinate, spherical harmonics (SH) coefficients to represent its color, an opacity parameter, a scale vector, and a rotation quaternion. 3D-GS utilizes scale vectors and rotation quaternions to characterize the covariance matrix of the 3D Gaussian function. The coordinates of Gaussians are typically referred to as \textit{geometry}, while the other parameters of Gaussians are denoted as \textit{attributes}.
Despite the efficiency of 3D-GS, the sheer volume of Gaussians and the multi-channel attributes within each Gaussian result in a considerable file size. Notably, $5.27 \times 10^6$ Gaussians are required to represent the \textit{bicycle} scene in the Mip-NeRF 360 dataset \cite{barron2022mip}, occupying 1.3 GB of storage under 32-bit float precision. This sizable file poses challenges in transmission and storage. Hence, it is essential to design a tailored codec for 3D Gaussians.

Due to the diverse attributes and intricate rendering procedure of the 3D-GS, developing an efficient compression method for 3D Gaussians presents significant challenges. Previous studies on point cloud compression \cite{zhang2014point, de2016compression, fang20223dac, song2023efficient} involve voxelizing the point cloud and applying transformations, quantization, and entropy encoding. However, these approaches cannot support fine-tuning to restore the quality of compressed 3D Gaussians and are limited to conventional point clouds with basic attributes like color and normals. In contrast, 3D-GS encompasses a broader spectrum of attributes. Blindly applying existing methods results in notable artifacts due to the high sensitivity of certain 3D-GS attributes. Hence, adapting traditional compression techniques for 3D Gaussians is non-trivial. Some concurrent works \cite{niedermayr2023compressed, fan2023lightgaussian, navaneet2023compact3d, girish2023eagles, morgenstern2023compact, jc2023gsplat, ap2023gssmall, ap2023gssmaller} have explored compressing 3D Gaussians using vector quantization and finetuning. 
Nevertheless, these approaches typically separate geometry from attributes during compression, disregarding the potential similarities among attributes at the 3D geometric level. This oversight prevents efficiently utilizing geometry information to further reduce attribute redundancy. Additionally, the requirement for training puts tremendous pressure on real-world applications.

To address the aforementioned issues, we introduce MesonGS, a 3D Gaussians codec, which employs universal Gaussian pruning, attribute transformation, and block quantization. 
In \emph{universal Gaussian pruning}, we consider both view-dependent and view-independent features to assess the importance of Gaussians.
The gradient-based importance \cite{niedermayr2023compressed} tends to select Gaussians with large gradients rather than Gaussians that contribute significantly to the rendering results, making it unsuitable for the poorly learned 3D-GS.
In contrast, our method accurately evaluates importance through forward propagation and is applicable in all scenarios.
Besides, compared to opacity-based importance \cite{fan2023lightgaussian}, we incorporate view-dependent color features to achieve more accurate evaluations.
Regarding \emph{attribute transformation}, we first transform the rotation quaternions (4 numbers) into Euler angles (3 numbers), a lossless process that reduces the storage requirement for each Gaussian by one number.
Then, we adopt region adaptive hierarchical transform (RAHT) \cite{de2016compression} to reduce the entropy of key attributes -- opacity, scales, Euler angles, and 0-degree SH coefficients.
RAHT involves transforming a channel of the attribute into a DC coefficient and several concentrated distributed AC coefficients. 
Since the entropy of AC coefficients is lower, the entropy coding methods can compress the attribute into a smaller size.
For \emph{block quantization}, we divide each attribute channel into multiple blocks and perform quantization for each block individually.
This approach prevents quantization from becoming the quality bottleneck and provides increased flexibility.
We employ vector quantization \cite{equitz1989new} to significantly compress unimportant attributes -- SH coefficients with degrees greater than 0. 
We utilize octree to compress geometry and pack all the elements with LZ77 \cite{gailly2003zlib, ziv1977universal, ziv1978compression} codec.
To achieve a fair comparison with related works and restore the quality, we also propose an elaborated finetune scheme.
Comprehensive experiments are conducted to demonstrate the exceptional compression quality of our method.
Additionally, we evaluate our method against previous neural radiance field (NeRF) compression techniques.

Our contributions can be summarized as follows:
\begin{itemize}
    \item We propose two transformations to reduce the redundancy and entropy in attributes. This involves using Euler angles to replace rotation quaternions and applying RAHT to key attributes. These transformations lead to a 13$\times$ increase
    in compression rate with negligible quality degradation.
    \item We derive a measure that considers both view-dependent and -independent factors to prune the insignificant Gaussians, suiting both bounded and unbounded scenes. We also adopt a finer-grained method to avoid excessive information loss caused by quantization. Finally, a well-crafted finetune scheme is devised to restore quality.
   \item Extensive experiments demonstrate the compression quality of our pipeline. 
    Comparisons with concurrent works demonstrate the universality of our pruning and transformation strategies. Additionally, we have compared our work with compressed NeRF to reveal the strengths and limitations between Grid-based NeRF and 3D Gaussians in the 1MB storage.

\end{itemize}
 
\section{Related Work}
\label{sec:rw}
\vspace{1mm}\noindent\textbf{3D Gaussians Compression.}
Many concurrent works are proposed to compress 3D Gaussians. C3DGS\cite{niedermayr2023compressed} proposes a score to measure the sensitivity of the attributes and replace the quaternions and scales with covariance. They use vector quantization to compress color and geometry features separately. Lee \etal \cite{lee2023compact} give a learning-based pruning strategy, utilizes residual vector quantization to compress the scales and rotations, and compresses the SH coefficients with a NeRF.
LightGaussian\cite{fan2023lightgaussian} compresses the geometry with an octree, prunes the Gaussians based on the cumulated opacity and volumes, distills the SHs to a lower degree, and finally compresses the remaining elements with vector quantization.
Compact3D\cite{navaneet2023compact3d} uses vector quantization and compresses the indices further by sorting them and using a method similar to run-length encoding. EAGLES\cite{girish2023eagles} treats the parameters of a Gaussian as a vector and proposes an encoder-decoder network to compress the vector into a latent code.
SOGS\cite{morgenstern2023compact} structures the Gaussian attributes into smooth 2D grids.
ReducingGS\cite{papantonakis2024i3d} introduces mixed-band SH coefficients to further reduce the file size.
These methods require a significant amount of training time to distill the information of original 3D Gaussians into the compact model, making them unsuitable for resource-constrained scenarios and inapplicable for 3D-GS extensions \cite{qin2023langsplat}.
Instead of relying on extensive training to identify redundancy in attributes for compression, we propose using RAHT to reveal spatial redundancy in attributes and then combining it with entropy coding to reduce the file size further. Besides, compared to \cite{fan2023lightgaussian, niedermayr2023compressed}, we propose a Gaussian importance assessment index that considers both view-dependent and -independent factors, suiting both bounded and unbounded scenes. Furthermore, we replace the quaternions with the Euler angles, showing better quality than covariance-based replacement \cite{niedermayr2023compressed}.

\vspace{1mm}\noindent\textbf{NeRF Compression.} 
Neural Radiance Field (NeRF) compression primarily targets grid-based NeRF \cite{chen2023factor,chen2023neurbf,barron2023zipnerf,hu2023Tri-MipRF}. 
The ``grid'' can be voxel-grids \cite{sun2022direct, DBLP:conf/cvpr/Fridovich-KeilY22}, tri-planes \cite{DBLP:conf/eccv/ChenXGYS22}, point clouds \cite{xu2022point}, or hash grids \cite{muller2022instant}. Though grid-based NeRF achieves great acceleration, it introduces huge storage requirements, leading to the emergence of NeRF compression works. VQAD \cite{DBLP:conf/siggraph/TakikawaET0MJF22} proposes to 
generalize different versions of a NeRF with hierarchical coding methods. \cite{deng2023compressing, zhao2023tinynerf, li2023compressing} propose to use frequency domain transformation to reduce the storage demand of the voxel grids.
SHACIRA \cite{girish2023shacira} and CAwa-NeRF \cite{mahmoud2023cawa} compress the hash grid of InstantNGP \cite{muller2022instant}. BiRF \cite{shin2023binary} introduces a binary quantization scheme.
Masked-wavelet-NeRF \cite{Rho_2023_CVPR} and ACRF \cite{fang2024acrf} adopt wavelet transform and RAHT.
ReRF \cite{wang2023neural} is proposed to compress the dynamic NeRF.
To accelerate rendering, some works \cite{DBLP:journals/tog/ReiserSVSMGBH23, DBLP:conf/iccv/HedmanSMBD21, DBLP:conf/cvpr/ChenFHT23} quantize features to 8 bits and save them as 2D images.
NeRF and 3D-GS can be interchanged in novel view synthesis, particularly on bounded scenes. Therefore, a comparison between them is valuable. Our work proposes such a comparison to reveal their core competitiveness.

\vspace{1mm}\noindent\textbf{Point Cloud Compression.}
Point cloud compression consists of geometry compression and attribute compression.
The goal of geometry compression is to compress the 3D coordinates of points. Existing methods \cite{meagher1982geometric,schwarz2018emerging}
typically use octree to organize coordinates. 
Attribute compression generally comprises three steps: transform coding, quantization, and entropy coding. 
Transform coding involves designing a careful transformation of attributes 
into the frequency domain to minimize signal redundancy. 
For instance, \cite{zhang2014point} constructs a graph from the point cloud and applies the graph fourier transform to attributes. 
\cite{de2016compression} introduces Haar wavelet transforms to attribute compression. Quantization is used to convert coefficients from transform coding into transmitted symbols
and reduce the high-frequency components. 
Entropy coding \cite{huffman1952method,witten1987arithmetic,weinberger2000loco,richardson2004h} aims to encode these symbols into a bitstream.
3DAC \cite{fang20223dac} and Song \etal \cite{song2023efficient} propose learning-based entropy models to further reduce the size. 

\section{Method}

\subsection{Preliminary}
3D-GS \cite{kerbl20233d} is an explicit 3D scene representation in the form of point clouds, utilizing Gaussians to model the scene. Each Gaussian is characterized by a covariance matrix $\mathbf{\Sigma}$ and a center point $\mathbf{X}$, which is referred to as the mean value of the Gaussian:
\begin{equation}
    \label{eq:gs_rep}
    G(x) = e^{-\frac{1}{2}\mathbf{X}^\top \mathbf{\Sigma}^{-1}\mathbf{X}}.
\end{equation}
To maintain the positive definiteness of the covariance matrix $\mathbf{\Sigma}$, 
3D-GS decomposes $\mathbf{\Sigma}$ into a scaling matrix $\mathbf{S} = {\rm diag}(\mathbf{s}), \mathbf{s} \in \mathbb{R}^3$ and a rotation matrix $\mathbf{R}$: $\mathbf{\Sigma} = \mathbf{R}\mathbf{S}\mathbf{S}^\top \mathbf{R}^\top$.
The rotation matrix $\mathbf{R}$ is parameterized by a rotation quaternion $\mathbf{q} \in \mathbb{R}^4$. The backpropagation process is illustrated in \cite{kerbl20233d}.

When rendering novel views, the technique of splatting \cite{ewa_volume_splatting,yifan2019differentiable} is employed for the Gaussians within the camera planes. As introduced by \cite{zwicker2001surface}, using a viewing transform denoted as $\mathbf{W}$ and the Jacobian $\mathbf{J}$ of the affine approximation of the projective transformation, the covariance matrix $\mathbf{\Sigma}'$ in camera coordinates system can be computed by $\mathbf{\Sigma}' = \mathbf{J} \mathbf{W} \mathbf{\Sigma} \mathbf{W}^\top \mathbf{J}^\top$.

In summary, each element of 3D Gaussians has the following parameters:
(1) a 3D center $\mathbf{\mu} \in \mathbb{R}^3$; (2) a rotation quaternion $\mathbf{q} \in \mathbb{R}^4$;
(3) a scale vector $\mathbf{s} \in \mathbb{R}^3$; 
(4) a color feature defined by spherical harmonics coefficients $\mathbf{SH} \in \mathbb{R}^d$, with $d = 3(f+1)^2$, where
$f$ is the harmonics degree; and (5) an opacity logit $o \in \mathbb{R}$. 
Specifically, for each pixel, the color and opacity of all the Gaussians are computed using \cref{eq:gs_rep}. 
The blending of $N$ ordered points that overlap the pixel is given by:
\begin{equation}
    \label{eq:alpha_comp}
    C = \sum_{i\in N}{c_i \alpha_i \prod_{j=1}^{i-1}(1 - \alpha_j)}.
\end{equation}
Here, $c_i$ and $\alpha_i$ represent the density and color of this point
computed by a Gaussian with covariance $\mathbf{\Sigma}$ multiplied by
an optimizable per-point opacity and SH color coefficients.

\subsection{Overview}
\begin{figure*}[t]
    \centering
     \includegraphics[width=0.99\linewidth]{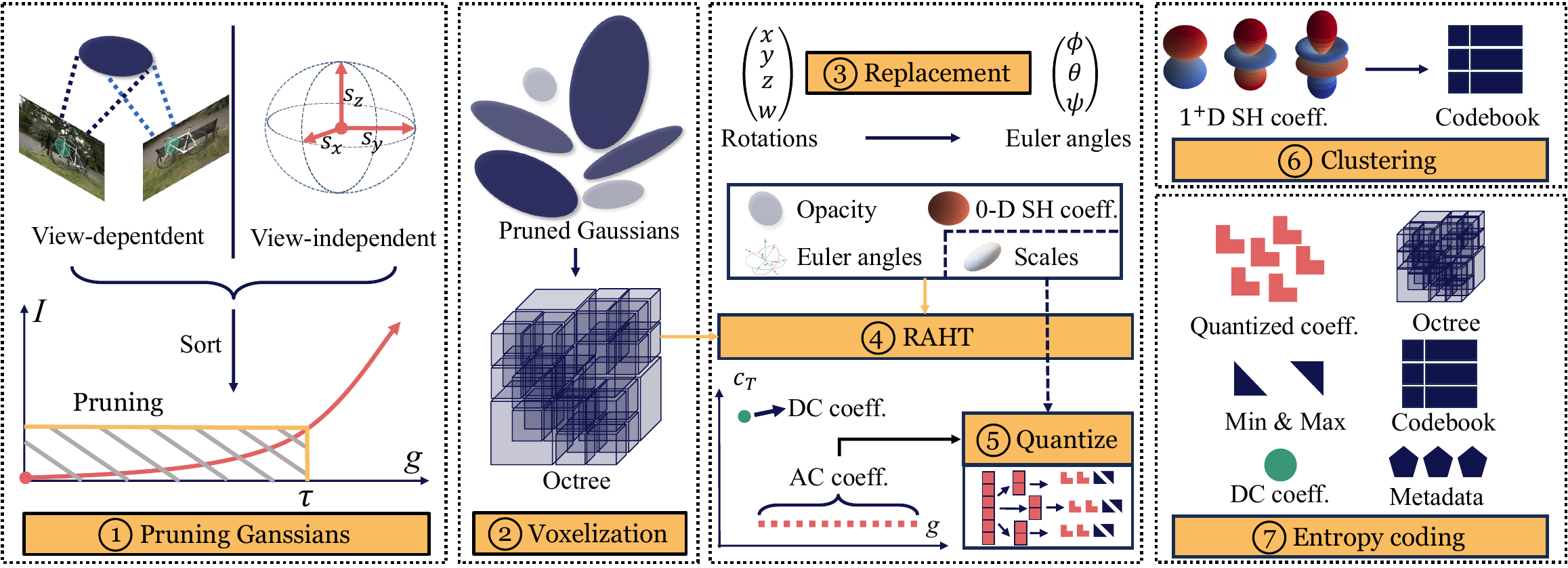}
     \caption{\textbf{Overview of MesonGS.} \textcircled{\raisebox{-0.9pt}{1}} We prune insignificant Gaussians by considering both view-dependent and view-independent features. \textcircled{\raisebox{-0.9pt}{2}} Geometry compression is performed using an octree to generate voxelized coordinates for future transformations. \textcircled{\raisebox{-0.9pt}{3}} We replace rotation quaternions with Euler angles. \textcircled{\raisebox{-0.9pt}{4}} Further compression is achieved by applying RAHT and \textcircled{\raisebox{-0.9pt}{5}} block quantization to important attributes. Notably, RAHT is not applied to the scales when the quantization bit is 8 (refer to the dashed arrow). \textcircled{\raisebox{-0.9pt}{6}} To significantly compress the $1^+$D SH coefficients, vector quantization is employed. \textcircled{\raisebox{-0.9pt}{7}} All components are packed by LZ77 \cite{gailly2003zlib, ziv1977universal, ziv1978compression} codec.}
     \label{fig:overview}
     \vspace{-0.5cm}
\end{figure*}

As shown in \cref{fig:overview}, we first prune insignificant 3D Gaussians based on view-dependent and -independent features. Then, we compress the remaining 3D Gaussians.
For geometric compression, octree is employed to compress the positions of the Gaussians.
For attribute compression, we begin by replacing rotation quaternions with Euler angles. Then, we categorize the attributes into important and unimportant ones, with the former including opacity, 0-D SH coefficients, scale vectors, and Euler angles and later including the SH coefficients in degrees greater than 0. We apply RAHT and block quantization to important attributes. Note that RAHT is not applied to the scale vectors at the 8-bit quantization. Unimportant attributes are significantly compressed through vector quantization. Finally, all components are packed using the LZ77 \cite{gailly2003zlib, ziv1977universal, ziv1978compression} codec.

\subsection{Gaussians Pruning}
As 3D-GS has a huge number of points, pruning unimportant Gaussians is a necessary step.
We first define the importance score of each Gaussian. Here the importance score refers to the contribution to the final rendering results.
For a Gaussian $g$, we define its importance score $I_g$ as the product of the view-dependent importance score $I_d$ and the view-independent importance score $I_i$: $I_g = I_d I_i$.
Based on \cref{eq:alpha_comp},
we define view-dependent importance score $I_d$ as: 
\begin{equation}
    I_d = \sum_{p \in \mathcal{P}} \alpha_i \prod_{j=1}^{i-1}(1 - \alpha_j).
    \label{eq:view_dep}
\end{equation}
Here, $\mathcal{P}$ is the pixel set that is overlapped by the Gaussian $g$,
and $i$ is the rank of Gaussian $g$ in a set of Gaussians that overlap with the pixel $p$.
In contrast to VQRF \cite{li2023compressing}, 
where the importance score is the mean value of corresponding sample points, our method allows for the direct recording of the importance score throughout the testing phase.
LightGaussian \cite{fan2023lightgaussian} uses the opacity ($\alpha_i$ in \cref{eq:view_dep}) as the view-dependent score. They have not considered the masking caused by other Gaussians. 
The usage of backward gradients as importance scores in C3DGS \cite{niedermayr2023compressed} is not ideal for 3D-GS that is not well-learned. Typically, Gaussians that are not well-learned exhibit larger gradients. However, these poorly learned Gaussians may not be the significant ones, making the pruning strategy of C3DGS ineffective in inadequately learned 3D Gaussians.

The view-independent score $I_i$ is given by $I_i = (V_{\rm norm})^{\beta}$.
Here, the volume $V$ is the product of the scale vector.
To obtain $V_{\rm norm}$, we normalize the $V$ by the 90\% largest
of all sorted Gaussians and clip the range between 0 and 1,
$\beta$ is the hyperparameter to control the size of $I_i$.

\begin{figure}[t]
    \centering
    \begin{subfigure}{0.48\linewidth}
        \includegraphics[width=0.99\linewidth]{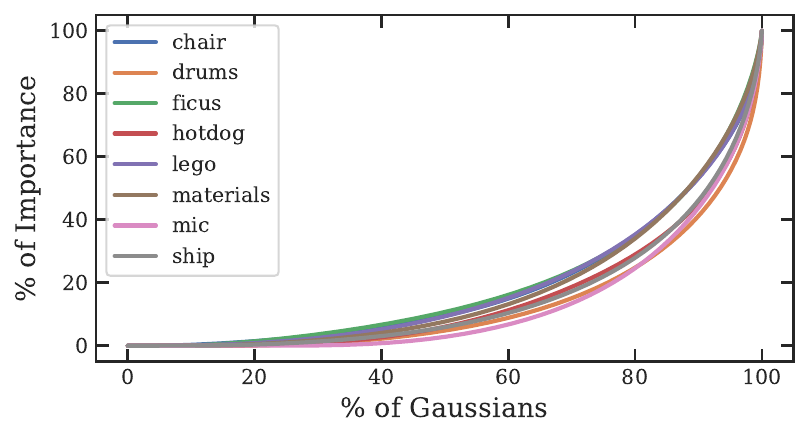}
      \caption{Bounded scenes.}
      \label{fig:syn_cdf}
    \end{subfigure}
    \hfill
    \begin{subfigure}{0.48\linewidth}
    \includegraphics[width=0.99\linewidth]{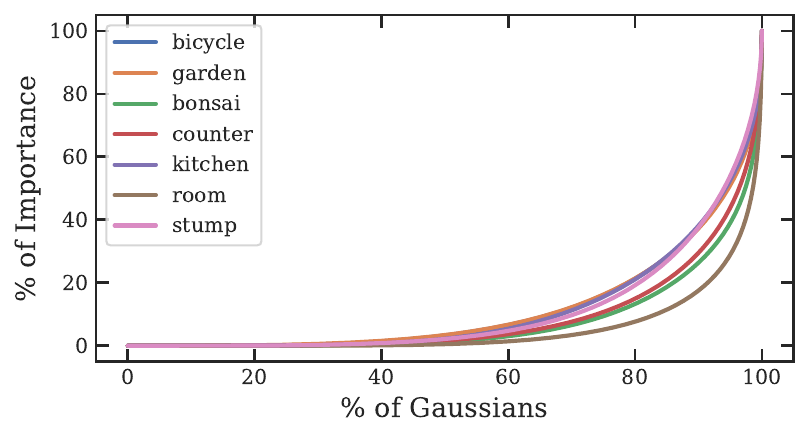}
      \caption{Unbounded scenes.}
      \label{fig:ub_cdf}
    \end{subfigure}
    \vspace{-0.2cm}
    \caption{\textbf{Quantile-quantile curve, which means $\mathbf{x}$\% of least important Guasisans
    contributes to $\mathbf{y}$\% percent of total importance.} 
    For both kinds of scenes, 40\% of the Gaussians contribute over 80\% of the importance.
    The importance refers to the contribution to the final rendering results.}
    \label{fig:cdf}
    \vspace{-0.5cm}
\end{figure}

In \cref{fig:cdf},
we sort the importance score and visualize its cumulative distribution function (CDF). We notice that 40\% of the Gaussians contain over 80\% of the importance. Hence, we use an importance threshold $\tau$ to prune Gaussians, meaning we cut the percent of $\tau$ of the sorted Gaussians.

\subsection{Geometry Compression}
After pruning, we compress the 3D positions with the octree structure. An octree recursively divides occupied voxels into eight sub-voxels until reaching the required resolution. The occupancy symbol is composed of 8 bits (1 to 255 in decimal), where each bit indicates the occupancy status of the corresponding subvoxel. We use the depth $d$ to control the size of the octree. When multiple Gaussians exist within a voxel, we average the corresponding attributes for deduplication.

\subsection{Attribute Transformation and Compression}
We categorize attributes into important attributes and unimportant attributes. Important attributes include opacity, scales, rotations/Euler angles, and 0D-SH coefficients. Here, the 0D-SH coefficient refers to SH coefficients in degree 0. The unimportant attributes refer to the SH coefficients in degrees greater than 0. 

\vspace{1mm}\noindent\textbf{Replacement.}
We replace the rotation quaternion (4 numbers) with the corresponding Euler angles (3 numbers).
The Euler angles are three angles that describe the orientation of a rigid body with respect to a fixed coordinate system. This replacement can reduce the storage requirement by one floating-point number for each Gaussian Point.

Specifically, for a quaternion $\mathbf{q} = [w, x, y, z] \in \mathbb{R}^4$, 
we calculate the euler angle $\mathbf{e} = [\phi, \theta, \psi] \in \mathbb{R}^3$ with:
\begin{equation}\scriptstyle
\left[ 
    \begin{array}{c}
        {\rm atan2}(2(wx + yz), 1 - 2(x^2 + y^2)) \\
        -\frac{\pi}{2} + 2{\rm atan2}(\sqrt{1 + 2(wy - xz)}, \sqrt{1 - 2(wy - xz)}) \\
        {\rm atan2}(2(wz+xy), 1 - 2(y^2 + z^2))
    \end{array}
\right].
\end{equation}
During decoding, we directly build the rotation matrix $\mathbf{R}$ from the Euler angles by:
\begin{equation}\scriptstyle
\left[ 
\begin{array}{ccc}
    {\rm C}_{\theta}{\rm C}_{\psi} & -{\rm C}_{\phi}{\rm S}_{\psi} + {\rm S}_{\phi}{\rm S}_{\theta}{\rm C}_{\psi} & {\rm S}_{\phi}{\rm S}_{\psi} + {\rm C}_{\phi}{\rm S}_{\theta}{\rm C}_{\psi} \\
    {\rm C}_{\theta}{\rm S}_{\psi} & {\rm C}_{\phi}{\rm C}_{\psi} + {\rm S}_{\phi}{\rm S}_{\theta}{\rm S}_{\psi} & -{\rm S}_{\phi}{\rm C}_{\psi} + {\rm C}_{\phi}{\rm S}_{\theta}{\rm S}_{\psi} \\
    -{\rm S}_{\theta} & {\rm S}_{\phi}{\rm C}_{\theta} & {\rm C}_{\phi}{\rm C}_{\theta} \\
\end{array}
\right].
\end{equation}
Here ${\rm S}_{\theta}$ and ${\rm C}_{\theta}$ represents sine and cosine of $\theta$. Similarly, $\phi$ and $\psi$ follow the same notation.

As the covariance matrix is symmetrical, replacing the scales and rotation quaternions (7 numbers) with the upper triangular part (6 numbers) of the covariance matrix seems to be an alternative.
This strategy can also reduce the storage of one number.
Our contemporary, C3DGS\cite{niedermayr2023compressed}, proposes a similar covariance-based replacement strategy. 
However, the subsequent quantization steps will cause a large number of covariance matrices to become indefinite, resulting in degraded rendering outcomes.
In contrast, using Euler angles can ensure that the positive definiteness of covariance is not compromised. The comparison results are illustrated in \cref{fig:euler_vs_cov}.

\begin{figure}[t]
    \centering
    \begin{minipage}[b]{0.31\textwidth}
        \centering
        \includegraphics[width=0.99\linewidth]{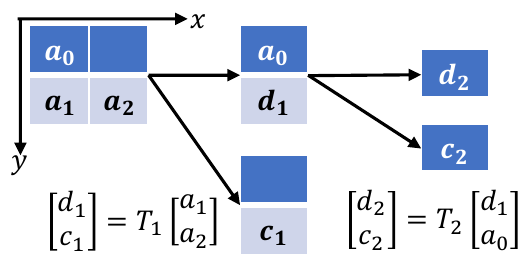}
        \caption{\textbf{2D example of RAHT.}}
        \label{fig:raht}
    \end{minipage}
    \begin{minipage}[b]{0.68\textwidth}
        \centering
        \includegraphics[width=0.99\linewidth]{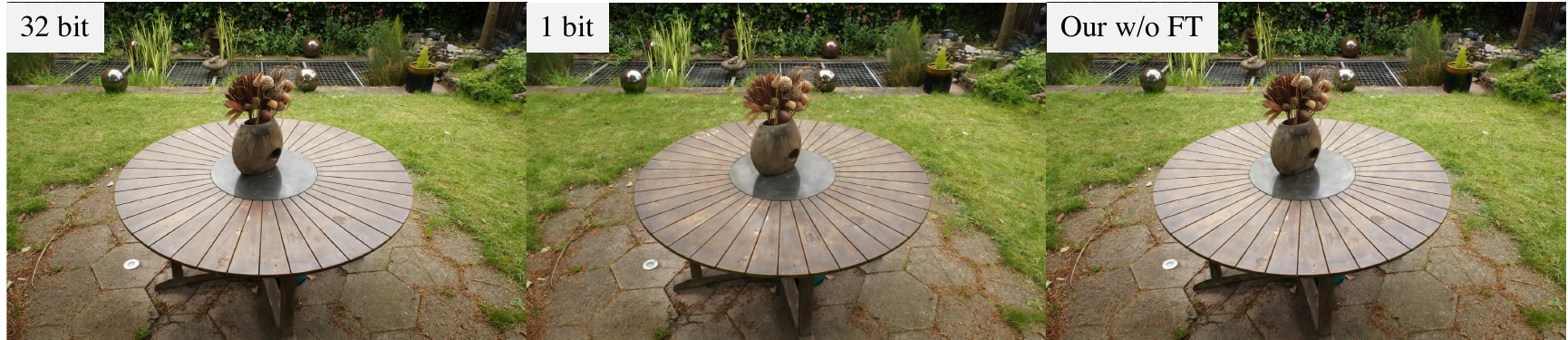}
        \caption{\textbf{Effect of $1^+$D spherical harmonics coefficients.} ``FT'' refers to finetune.}
        \label{fig:cluster}
    \end{minipage}
    \vspace{-1cm}
\end{figure}

\vspace{1mm}\noindent\textbf{Region adaptive hierarchical transform.}
To reduce signal redundancy in important attributes, 
we employ RAHT \cite{de2016compression}.
RAHT involves taking voxelized coordinates from octree
and converting the corresponding attributes into transformed coefficients. 
Each channel of the transformed coefficients consists of a direct current (DC) coefficient
and several alternating current (AC) coefficients.
The low entropy of the AC coefficients enables the subsequent entropy coding procedure to achieve a larger compression rate. 
Here, we briefly introduce the RAHT through a 2D example.
\cref{fig:raht} shows how to apply RAHT to $a_0$, $a_1$, and $a_2$.
First, RAHT merges the coefficients along the $x$ axis with the transform:
\begin{equation}
    T_1 = \frac{1}{\sqrt{w_1 + w_2}} 
    \left[
        \begin{array}{cc}
        \sqrt{w_1} & \sqrt{w_2} \\
        -\sqrt{w_2} & \sqrt{w_1} \\
        \end{array}
    \right],
\end{equation}
where the weight coefficient $w_i$ is the number of Gaussians that $a_i$ contains. 
If $a_i$ is a leaf node in octree, then $w_i = 1$. If $a_i$ is not a leaf node,
then $w_i$ is the sum weight of its son leaf nodes.
After applying $T_1$ to $a_1$ and $a_2$, we get a DC coefficient $d_1$ and 
an AC coefficient $c_1$. DC coefficient $d_1$ is going to do further transformation with $a_0$ while AC coefficients are saved for encoding.
For coefficients that have no counterpart, like $a_0$, we transmit it to the next layer.
At the end, we obtain a DC coefficient $d_2$ and two AC coefficients $c_2$ and $c_1$.
During the decoding process, we obtain the weight coefficients from the octree.

\vspace{1mm}\noindent\textbf{Block quantization.}
After applying RAHT to important attributes, we save the DC coefficient in float and quantize the AC coefficients.
We use block-wise quantization \cite{frantar-gptq,2023-emnlp-osplus,MLSYS2024_5edb57c0,tang2022mixed} to prevent significant quality degradation caused by the coarse-grained channel-wise quantization.
Specifically, we first partition a channel of attributes into multiple blocks. Then we quantize a block $\mathbf{c}$ with:
\begin{equation}
    \mathbf{c}_q = \lfloor {\rm clamp}(\frac{\mathbf{c}}{S_c} + Z_c, 0, 2^b - 1) \rceil,
\end{equation}
where
\begin{equation}
    S_c = \frac{\max(\mathbf{c}) - \min(\mathbf{c})}{2^{b}},\space\space Z_c = \lfloor 2^b - \frac{\max(\mathbf{c})}{S_c} \rceil.
\end{equation}
Here, $b$ refers to the bit-width, $\lfloor \cdot \rceil$ represents the rounding-to-nearest function, and $\mathbf{c}_q$ refers to the quantized attributes. Besides, function ${\rm clamp}(\cdot)$ specifies a range of values. Values below the minimum are set to the minimum. Values above the maximum are set to the maximum. 

Note that we do not apply RAHT to the scale vectors when the quantization bit is 8.
The reason is that the activation function of the scale vector is an exponential function, which 
magnifies the errors caused by transformation and quantization.
We save the minimum and maximum values of all blocks of quantized attributes for decoding.

\vspace{1mm}\noindent\textbf{Clustering $1^+$D SH coefficients.}
The size of spherical harmonics (SH) coefficients takes up 85.7\% in a 3D Gaussians file.
In the middle of \cref{fig:cluster}, after applying 1-bit quantization to the $1^+$D SH coefficients, only the reflectance is affected, while the overall structure and color remain unchanged. 
Therefore, quantization is not the optimal choice for $1^+$D SH coefficients.
We employ vector quantization to significantly reduce the size of these SH coefficients.
Specifically, we use a codebook and a corresponding index mapping table to 
connect the origin vectors with the vectors in the codebook.
The right side of \cref{fig:cluster} shows the final result of our method.
To reduce memory occupation and encoding time, a batched clustering strategy \cite{web_scale_cluster} is used.
We use multiple iterations to update the clustering results.

\vspace{1mm}\noindent\textbf{Encoding.}
The final file contains the following components:
(1) Octree; 
(2) DC coefficients and Quantized coefficients; 
(3) Codebook and the corresponding mapping table; 
(4) Metadata: Min-Max values of each block of quantized coefficients, octree depth, block size.
Notably, the DC coefficients, codebook, and metadata are stored in floating-point format, whereas other components are saved as integers. We compress them via LZ77\cite{gailly2003zlib, ziv1977universal, ziv1978compression} codec.

\vspace{1mm}\noindent\textbf{Finetune.}
We propose a finetune scheme to achieve a fair comparison with baselines and solve the problem of former methods not supporting backpropagation.
Specifically, we fix the coordinates of pruned 3D Gaussians and only finetune the attributes.
We simulate the encoding and decoding processes during the forward process.
To pass the gradients during the backward process, 
we employ the straight-through estimator \cite{bengio2013estimating} for quantization.

\section{Experiments} 
\label{sec:exp}

\vspace{1mm}\noindent\textbf{Datasets and compression settings.} 
(1) \textbf{Mip-NeRF 360.} The Mip-NeRF 360 dataset \cite{barron2022mip} contains five outdoor and four indoor scenes. 
Each scene contains 100 to 300 images. We use the images at 1600$\times$1063. Note that the undocumented \textit{flower} and \textit{treehill} scenes are not included in our evaluation.
(2) \textbf{Tank\&Temples.} This dataset \cite{tandt2017} contains two scenes, including \textit{train} and \textit{truck}.
(3) \textbf{Deep Blending.} This dataset \cite{hedman2018db} contains two scenes, including \textit{drjohnson} and \textit{playroom}.
(4) \textbf{Synthetic-NeRF.} This dataset was first introduced by \cite{mildenhall2021nerf} and has been widely adopted by subsequent work. It contains 8 scenes rendered at 800$\times$800 by Blender. Each scene contains 100 rendered views as the training set and
200 views for testing. As for train-test split and camera parameters estimation, we follow the official implementation of 3D-GS. 
We evaluate rendering quality and compression performance using PSNR, 
SSIM \cite{wang2004image}, LPIPS \cite{zhang2018unreasonable}, size, compression rate, and decoding time. All the metrics are evaluated on the test set if not otherwise specified. 
We set the bit-width as 8 if not otherwise specified. 
When calculating the important score, we only use the training dataset. To obtain the pre-trained 3D Gaussians for compression, we train $30,000$ iterations and then save the checkpoints for both datasets. We set the background as white. 
Please check the supplementary material for more details.

\begin{table}[t]
  \centering
  \caption{\textbf{Quantitative comparison on Mip-NeRF 360, Tank\&Temples, and Deep Blending.} The best results overall are bolded in each metric, and the second-best results are underlined.}
  \vspace{-0.2cm}
  \resizebox{\textwidth}{!}{
  \begin{tabular}{@{}l|cccc|cccc|cccc@{}}
  \hline
  \multirow{2}{*}{Method} & \multicolumn{4}{c|}{Mip-NeRF 360} & \multicolumn{4}{c|}{Tank\&Temples}       & \multicolumn{4}{c}{Deep Blending} \\
  \cline{2-13}
                          & PSNR   & SSIM   & LPIPS & Size (M)   & PSNR  & SSIM  & LPIPS & Size (M)   & PSNR   & SSIM   & LPIPS  & Size (M)   \\
  \hline
  3DGS   \cite{kerbl20233d}     & 28.98  & 0.865  & 0.193 & 641.70 & 23.36 & 0.838 & 0.187 & 421.90 & 29.56  & 0.898  & 0.250  & 703.77 \\
  C3DGS \cite{niedermayr2023compressed}  & 28.49  & \textbf{0.858}  & \textbf{0.205} & \underline{27.82}  & \textbf{23.32} & \underline{0.832} & \underline{0.194} & \underline{17.28}  & 29.38  & 0.898  & \textbf{0.238}  & \underline{25.30}  \\
  Lee \etal   \cite{lee2023compact}            & \underline{28.60}  & \underline{0.856}  & 0.209 & 46.98  & \textbf{23.32} & 0.831 & 0.201 & 39.40  & \textbf{29.79}  & \textbf{0.901}  & 0.258  & 43.20  \\
  \hline
  Our                     & 27.70  & 0.838  & 0.224 & \textbf{27.62}  & \underline{22.85} & 0.822 & 0.208 & \textbf{16.99}  & 29.08  & 0.895  & 0.260  & \textbf{24.76}  \\
  Our-FT                  & \textbf{28.61}  & \underline{0.856}  & \underline{0.206} & \textbf{27.62}  & \textbf{23.32} & \textbf{0.837} & \textbf{0.193} & \textbf{16.99}  & \underline{29.51}  & \textbf{0.901}  & \underline{0.251}  & \textbf{24.76} \\
  \hline
  \end{tabular}
  }
  \label{tab:mip360-t}
\end{table}

\begin{table}[t]
  \centering
  \caption{\textbf{Quantitative comparison on Synthetic-NeRF.} The best results overall are bolded in each metric, and the second-best results are underlined.}
  \vspace{-0.2cm}
  \begin{tabular}{@{}l|cccc||l|cccc@{}}
  \hline
  Method    & PSNR  & SSIM  & LPIPS & Size (M)  & Method & PSNR  & SSIM  & LPIPS & Size (M)       \\
  \hline
  3DGS  \cite{kerbl20233d}      & 33.37 & 0.970 & 0.031 & 68.55 & DVGO \cite{sun2022direct}  & 31.90 & 0.956 & 0.035 & 105.92     \\
  C3DGS  \cite{niedermayr2023compressed}     & \underline{32.94} & \underline{0.967} & \textbf{0.033} & 3.68  & VQRF \cite{li2023compressing}  & \underline{31.77} & \underline{0.954} & \textbf{0.036} & 1.43       \\
  Lee \etal \cite{lee2023compact} & \textbf{33.33} & \textbf{0.968} & \underline{0.034} & 8.61  & ACRF \cite{fang2024acrf}  & \textbf{31.79} & \underline{0.954} & \underline{0.037} & \underline{1.15}       \\
  \hline
  Our     & 32.25 & 0.963 & 0.038 & \textbf{3.65}  & Our  & 29.37 & 0.947 & 0.051 & \textbf{1.03} \\
  Our-FT    & 32.92 & \textbf{0.968} & \textbf{0.033} & \underline{3.66}  & Our-FT & 31.75 & \textbf{0.962} & 0.042 & \textbf{1.03} \\
  \hline
  \end{tabular}
  \label{tab:syn_main}
  \vspace{-0.5cm}
\end{table}

\begin{figure*}[t]
    \centering
    \begin{subfigure}{0.99\linewidth}
      \includegraphics[width=0.99\linewidth]{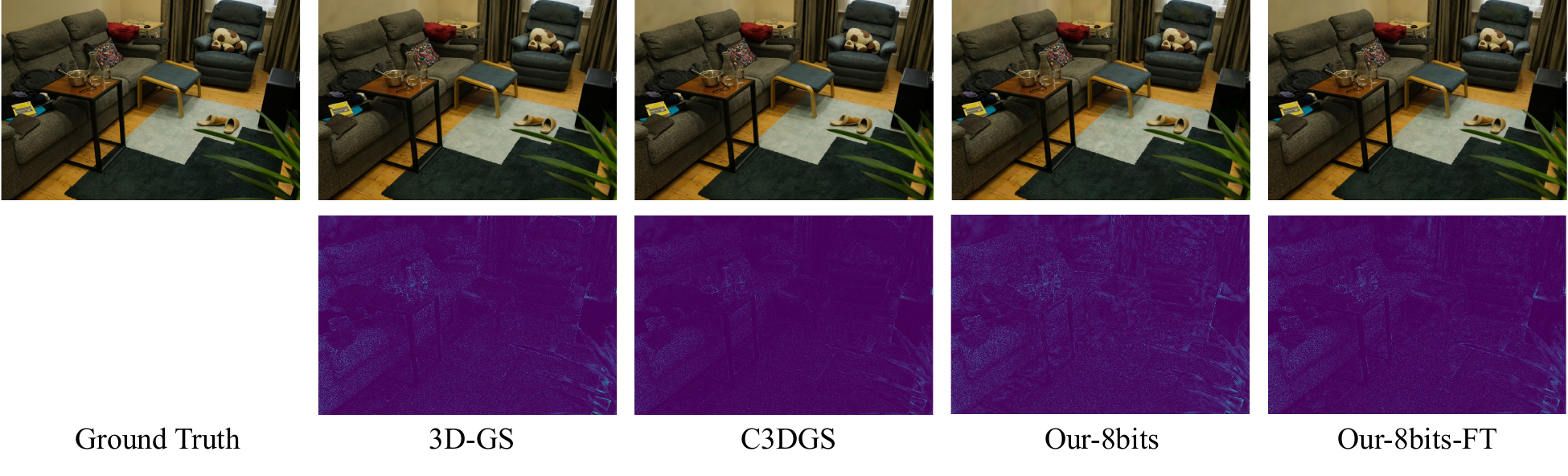}
      \vspace*{-5pt}    
      \caption{The \textit{Room} scene in the Mip-NeRF 360 dataset.}
        \label{fig:ub_main}
      \end{subfigure}
    \hfill
    \begin{subfigure}{0.99\linewidth}
    \includegraphics[width=0.99\linewidth]{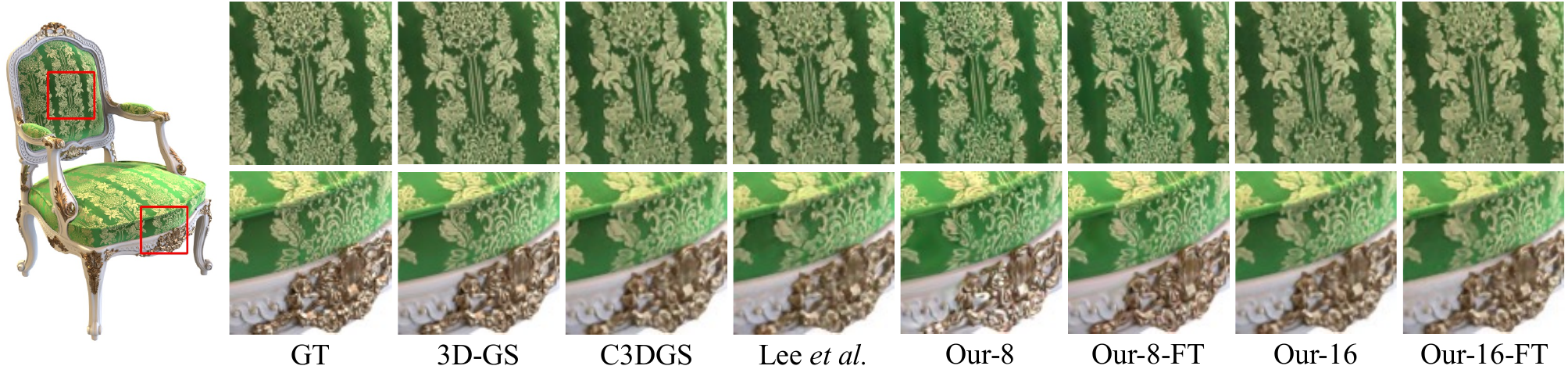}
    \vspace*{-5pt}  
    \caption{The \textit{Chair} scene in the Synthetic-NeRF dataset.}
      \label{fig:syn_main}
    \end{subfigure}
    \vspace*{-5pt}  
    \caption{\textbf{Qualitative comparison.} We can hardly observe visual artifacts on the rendering result of the compressed model compared to its original model.}
    \label{fig:vis_main}
\end{figure*}

\begin{figure}[t]
  \centering
  \vspace{-0.3cm}
  \includegraphics[width=\linewidth]{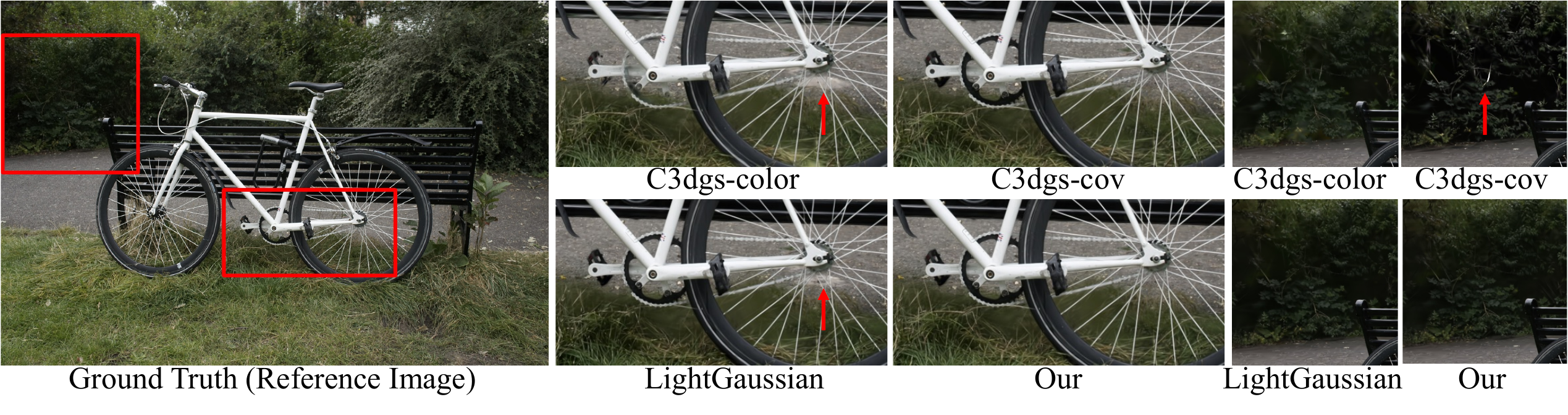}
  \vspace{-0.8cm}
  \caption{\textbf{Qualitative comparison on pruning strategy.} The visual quality of our pruning strategy is better than others.}
  \label{fig:r3q1}
  \vspace{-0.6cm}
\end{figure}

\vspace{1mm}\noindent\textbf{Baselines.} 
We compare our method with the concurrent 3D Gaussian compression works \cite{niedermayr2023compressed,lee2023compact} and NeRF compression works \cite{li2023compressing,fang2024acrf}. We use the evaluation results from their papers \cite{niedermayr2023compressed,lee2023compact,fang2024acrf}. 

\subsection{Experimental Results}
\vspace{1mm}\noindent\textbf{Quantitative result.} We compared our approach with original 3D-GS and concurrent works. 
All of the model sizes are calculated after a standard zip compression. 
As shown in \cref{tab:mip360-t} and \cref{tab:syn_main}, MesonGS realizes satisfactory performance across all the combinations of methods and datasets. The offline version of MesonGS can achieve a quality similar to the baselines. Moreover, after finetuning, MesonGS achieves performance comparable to the baselines, which reveals the efficiency of our finetune scheme.

\vspace{1mm}\noindent\textbf{Comparisons with NeRF compression.} By now, there is no work comparing compressed 3D Gaussians with compressed NeRF. 
The right side of \cref{tab:syn_main} shows the comparison of our 3D Gaussian compression method on Synthetic-NeRF with ACRF and VQRF.
We can observe that the quality metrics of baselines are slightly better.
The reason is two-fold. On the one hand, our finetune scheme cannot support updating the coordinates. 
If appropriate updates could be made to the coordinates, 
the overall quality of our method will be improved. 
On the other hand, the file sizes of baselines are larger.

\vspace{1mm}\noindent\textbf{Visualization result.} 
We compare the rendering results of  all baselines in \cref{fig:vis_main}
across bounded and unbounded scenes.
The visual difference is hard to observe in both \textit{chair} and \textit{room} scenes.
We also display the per-pixel mean absolute error between ground truth and other baselines at the second line.
Our finetune scheme efficiently restores the quality loss caused by offline compression and is highly consistent with the rendering results of 3D-GS.

\vspace{1mm}\noindent\textbf{Pruning strategy.} Here we compare our pruning strategies with LightGaussian \cite{fan2023lightgaussian} and C3DGS \cite{niedermayr2023compressed}. To achieve a fair comparison,
we prune 66\% of the Gaussians for all methods. \cref{tab:prune} shows
the superiority of our pruning strategy.
Moreover, \cref{tab:prune} indicates that it is necessary to 
incorporate view-dependent importance score $I_d$ in the pruning procedure.
Note that the output sizes of these strategies are the same because 
they prune the same percentage of Gaussians.
We also provide a qualitative comparison in Fig.~\ref{fig:r3q1} and use the red arrows and circles to mark the artifacts.

\begin{table}[t]\footnotesize
  \centering
  \caption{\textbf{Quantitative comparison on pruning strategy.} Our pruning strategy performs better on average. The best results overall are bolded in each metric, and the second-best results are underlined.}
  \vspace{-0.3cm}
  \label{tab:prune}
  \begin{tabular}{@{}l|ccc|ccc|c@{}}
    \hline
    \multirow{2}{*}{Methods} 
    & \multicolumn{3}{c|}{Synthetic-NeRF}  & \multicolumn{3}{c|}{Mip-NeRF 360 } & \multirow{2}{*}{Average PSNR} \\ 
    \cline{2-7} 
    & \multicolumn{1}{c}{PSNR} & \multicolumn{1}{c}{SSIM} & \multicolumn{1}{c|}{LPIPS}
    & \multicolumn{1}{c}{PSNR} & \multicolumn{1}{c}{SSIM} & \multicolumn{1}{c|}{LPIPS} \\
    \hline
    C3DGS-color \cite{niedermayr2023compressed} & \textbf{30.79} & \textbf{0.9606} & \textbf{0.0375} & 22.89 & 0.7852 & 0.2471 & 26.84 \\
    C3DGS-cov \cite{niedermayr2023compressed} & 24.37 & 0.9104 & 0.0738 & 15.04 & 0.6992 & 0.3044 & 19.71 \\
    LightGaussian \cite{fan2023lightgaussian} & 26.75 & 0.9372 & 0.0568 & \underline{26.93} & \underline{0.8327} & \underline{0.2290} & 26.84 \\
    \hline
    Our($I_i$) & 30.39 & 0.9591 & 0.0389 & 27.01 &  0.8411 & 0.2164 & \underline{28.70} \\
    Our($I_iI_d$) & \underline{30.52} & \underline{0.9597} & \underline{0.0383} & \textbf{27.52} & \textbf{0.8441} & \textbf{0.2139} & \textbf{29.02} \\
    \hline
  \end{tabular}
  \vspace{-0.3cm}
\end{table}

\vspace{1mm}\noindent\textbf{Encoding time.} As shown in Tab.~\ref{tab:enc_time}, without finetuning, MesonGS can complete compression in only 20\% of the time compared to concurrent works while maintaining similar rendering quality. When the number of Gaussian points is less than 20,000, MesonGS can complete compression quickly using only the CPU, while baseline methods require GPU assistance. 

\begin{table}[t]
    \centering
    \caption{\textbf{Encoding time.} Our method is the fastest.}
    \vspace{-0.3cm}
    \resizebox{\textwidth}{!}{
    \begin{tabular}{@{}l|ccc|ccc@{}}
    \hline
    \multirow{2}{*}{Method} & \multicolumn{3}{c|}{NeRF-Synthetic} & \multicolumn{3}{c}{MipNeRF-360} \\
    \cline{2-7}
                              &C3DGS \cite{niedermayr2023compressed}   & Lee \etal \cite{lee2023compact}  & MesonGS   & C3DGS \cite{niedermayr2023compressed}  & Lee \etal \cite{lee2023compact} & MesonGS  \\
    \hline
    Encoding Time            & 30 s     & 480 s       & \textbf{4 s}       & 5 min   & 33 min     & \textbf{1 min}  \\
    \hline
    \end{tabular}
    }
    \vspace{-0.5cm}
    \label{tab:enc_time}
\end{table}

\vspace{1mm}\noindent\textbf{Composition of final storage.}
In the Synthetic-NeRF dataset, the proportions of the octree, metadata, important attributes, and unimportant attributes are 43\%, 0.04\%, 34\%, and 23\%, respectively. In the Mip-NeRF 360 dataset, the proportions of the above four elements are 39\%, 0.02\%, 47\%, and 14\%, respectively. The octree and important attributes take up more than half of the storage, while metadata occupies a tiny proportion. The hyperparameter setting is consistent with the results in \cref{tab:syn_main} and \cref{fig:ub_main}.

\begin{figure}[t]
  \centering
  \vspace{-10pt}
   \includegraphics[width=0.99\linewidth]{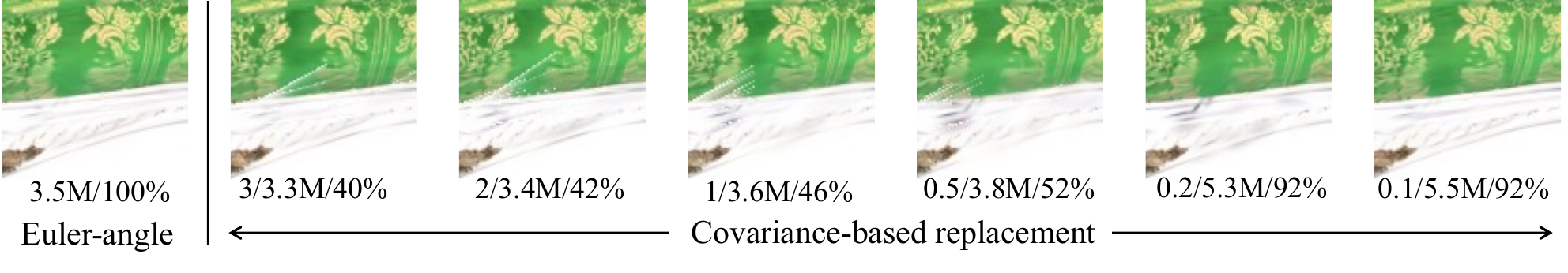}
   \vspace{-0.2cm}
   \caption{\textbf{Euler-angle-based vs. Covariance-based.} 
   ``A/B/C'' refers to the ``$\lambda_{\mathbf{c}}$ / the size of the compressed file / the percent of positive-definite covariance matrices''. Replacing scales and rotations with covariance leads to white line artifacts, which greatly affects the visual effect.
     We adjust the final file size by compressing a portion of the covariance using the $\lambda_{\mathbf{c}}$.}
   \label{fig:euler_vs_cov}
   \vspace{-0.2cm}
\end{figure}

\vspace{1mm}\noindent\textbf{Replacement strategy.} 
We set the bit width as 16 and show the close-up rendering results of Euler-angle-based replacement and 
covariance-based replacement in the left of \cref{fig:euler_vs_cov}. 
We adjust the final file size by compressing a portion of the covariance.
We can see some white line artifacts on the right of the covariance-based strategy.
The reason is that the lots of covariance matrices are not 
positive definite after the quantization, i.e., 92\% for Euler angle-based vs. $\sim$50\% for covariance-based replacement.
Please find more discussion in the supplementary material.

\begin{figure}[t]
  \centering
  \includegraphics[width=0.99\linewidth]{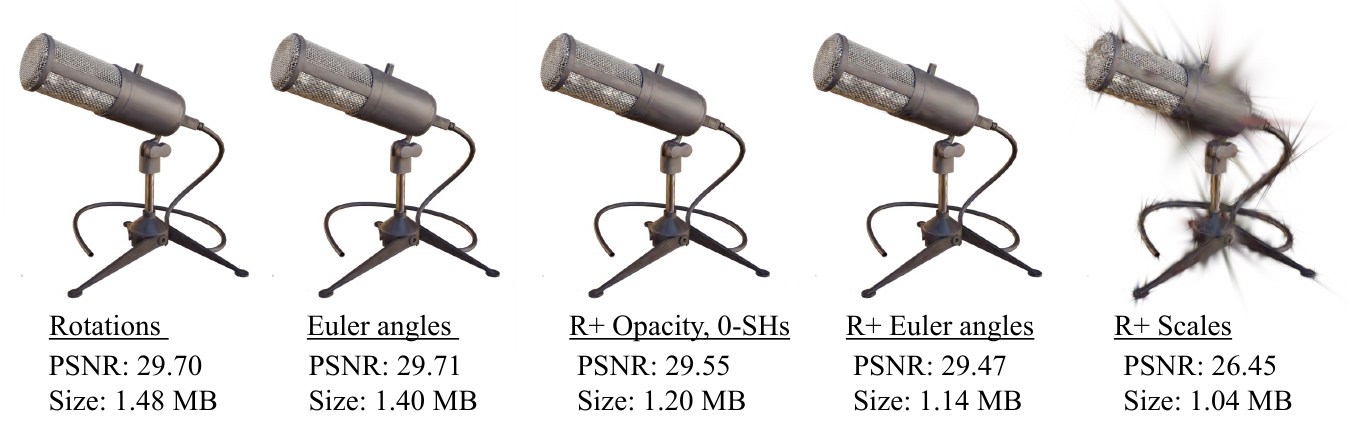}
  \vspace{-0.3cm}
  \caption{\textbf{Visual results of different attribute transformation stages.} 
  The first image from the left shows the baseline, which means saving the rotations quaternions in the final storage. 
  The second image shows the rendering results after replacing Rotation quaternions with Euler angles.
  ``R+*'' refers to applying RAHT to *.}
  \label{fig:raht_ab}
   \vspace{-0.2cm}
\end{figure}

\vspace{1mm}\noindent\textbf{RAHT and quantization.} For 8-bit quantization, we recommend not to apply RAHT for scales. Due to the activation function of the scale being an exponential function, it is more sensitive than other attributes.
The empirical evidences are shown in \cref{fig:raht_ab}.
After sequentially applying the RAHT to Opacity, 0D-SH coefficients, and Euler angles, we observe minor alterations in the rendering quality. However, severe degradation in rendering quality occurs after applying RAHT to Scales.
The information loss caused by the operation of RAHT + Quantization 
is more significant than only using Quantization, 
and the exponential function amplifies this error, 
leading to severe performance degradation.

\vspace{1mm}\noindent\textbf{Block quantization.} 
As shown in \cref{tab:imp}, when employing the per-channel quantization strategy and lowering the pruning threshold from 66\% to 50\%, the compressed file size expands while the rendering quality diminishes. This decline in quality stems from the amplified information loss linked to the extended channel length. As the number of Gaussian points rises, the information loss further intensifies. In contrast, \textit{block quantization} fixes the size of the quantization unit, thereby reducing information loss while providing more flexibility.

\begin{table}[t]\footnotesize
\center
\caption{\textbf{The advantage of block quantization.}
    By fixing the length of the vector requiring quantization, block quantization prevents quantization from becoming a performance bottleneck and provides more flexibility.}
  \label{tab:imp}
  \vspace{-0.2cm}
  \begin{tabular}{@{}l|c|cccc|cccc@{}}
  \hline
\multirow{2}{*}{Strategy} & \multirow{2}{*}{$\tau$} & \multicolumn{4}{c|}{Synthetic-NeRF}      & \multicolumn{4}{c}{Mip-NeRF 360}        \\
\cline{3-10}
                          &                            & PSNR(dB) & SSIM   & LPIPS  & Size(MB) & PSNR(dB) & SSIM   & LPIPS  & Size(MB) \\
\hline
\multirow{2}{*}{Channel}  & 66\%                       & 29.47     & 0.9476 & 0.0511 & 1.14      & 25.30     & 0.7533 & 0.3074 & 11.64     \\
                          & 50\%                       & 30.65     & 0.9529 & 0.0475 & 1.59      & 25.35     & 0.7461 & 0.3147 & 16.47     \\
                          \hline
\multirow{2}{*}{Block}    & 66\%                       &  29.60  &    0.9494    &    0.0490	     &    1.21      & 26.28     & 0.8035 & 0.2598 & 12.46     \\
                          & 50\%                       & 30.97     & 0.9560 & 0.0441 & 1.73      & 27.20     & 0.8238 & 0.2402 & 18.42  \\   
\hline
\end{tabular}
 \vspace{-0.3cm}
\end{table}

\subsection{Ablation Study}
We conducted several ablation studies on the Mip-NeRF 360 dataset
and the Synthetic-NeRF dataset. 
If not otherwise specified, all the experimental results below have not undergone finetuning.
Please find more ablation studies in the supplementary material.

\vspace{1mm}\noindent\textbf{Performance of different stages.}
We conduct an experiment to demonstrate the benefit of each module in MesonGS. 
We calculate the size after a zip compression for fair comparison. 
As shown in \cref{tab:diffstage}, compared to the uncompressed baseline, Pruning achieves 
5$\times$ compression but causes a significant PSNR drop for bounded 360 scenes.
However, such a drop is slighter for unbounded scenes.
The following stages are all influenced by the pruning stage.
Of course, all of these stages, instead of Replacement, have caused 
varying degrees of damage to the rendering quality.
We can see that the Replacement step is indeed a free lunch for the 3D-GS attribute compression.

\begin{table}[t]\footnotesize
  \center
  \caption{\textbf{Ablation study of different stages.} ``TQ'' refers to applying RAHT and quantization on important attributes. ``Q-scales'' refers to quantizing scales.}
   \vspace{-0.5cm}
    \label{tab:diffstage}
  \begin{tabular}{@{}l|cccc|cccc@{}}
  \hline
  \multirow{2}{*}{Stages} & \multicolumn{4}{c|}{Synthetic-NeRF} & \multicolumn{4}{c@{}}{Mip-NeRF 360} \\
  \cline{2-9} 
  & \multicolumn{1}{c}{PSNR (dB)} & \multicolumn{1}{c}{SSIM} & \multicolumn{1}{c}{LPIPS} & \multicolumn{1}{c|}{Size (MB)}
  & \multicolumn{1}{c}{PSNR (dB)} & \multicolumn{1}{c}{SSIM} & \multicolumn{1}{c}{LPIPS}& \multicolumn{1}{c@{}}{Size (MB)} \\
  \hline
    3D-GS                & 33.37     & 0.9696 & 0.0305 & 68.55     & 28.98     & 0.8647 & 0.1931 & 641.73    \\
+Prune                  & 30.52     & 0.9597 & 0.0383 & 20.99     & 28.69     & 0.8612 & 0.1970 & 265.94    \\
+Voxel                  & 30.44     & 0.9592 & 0.0388 & 20.72     & 28.68     & 0.8610 & 0.1971 & 260.58    \\
+Replace                & 30.44     & 0.9592 & 0.0388 & 20.36     & 28.68     & 0.8610 & 0.1971 & 255.61    \\
+Cluster                & 29.71     & 0.9513 & 0.0469 & 5.63      & 27.74     & 0.8427 & 0.2187 & 79.35     \\
+TQ                     & 29.37     & 0.9476 & 0.0510 & 1.95     & 27.20     & 0.8239 & 0.2401 & 30.68     \\
+Q-scales               & 29.37     & 0.9474 & 0.0511 & 1.03      & 27.20     & 0.8238 & 0.2402 & 18.43     \\
+Fine-tune               & 31.75     & 0.9618 & 0.0416 & 1.03      & 27.45     & 0.8273 & 0.2357 & 18.40    \\    
  \hline  
\end{tabular}
 \vspace{-0.5cm}
\end{table}

\vspace{1mm}\noindent\textbf{Importance threshold $\tau$.} 
Pruning is a necessary step for 3D Gaussian compression. In \cref{tab:imp}, under the bit width of 8, we observe that pruning 33\% of points resulted in lower performance compared to pruning 66\% of points. The reason for this counterintuitive performance drop is that the bit depth of the quantization step is too low, and pruning 33\% of points in the 3D-GS model of unbounded scenes still leaves too many points, resulting in excessive information loss after quantization. Therefore, the quantization bit-width is the bottleneck of overall performance here. In \cref{tab:imp}, after increasing the quantization bit-width to 16, the performance of the 33\% surpasses other baselines.

\section{Conclusion}
\label{sec:conc}
In this paper, we propose an elaborated designed 3D Gaussians codec. We propose several key components, including the universal Gaussian pruning strategy, elaborated attribute transformation, and flexible block quantization. Extensive experiments demonstrate the superior performance of MesonGS.

\subsection*{Acknowledgments}
This work is supported in part by National Key Research and Development Project of China (Grant No. 2023YFF0905502) and Shenzhen Science and Technology Program (Grant No. JCYJ20220818101014030). We thank anonymous reviewers for their valuable advice and JiangXingAI for sponsoring the research.

\bibliographystyle{splncs04}
\bibliography{main}
\end{document}